\documentclass{article}
\usepackage[preprint]{packages/neurips2023}
\usepackage{packages/extra_packages}

\title{Less is More! \\ A slim architecture for optimal language translation}

\date{ }

\author{
Luca Herranz-Celotti$^{1}$ \quad Ermal Rrapaj$^{2}$ \\
$^1$NECOTIS, Universit\'e de Sherbrooke, Canada \quad \\ $^2$ NERSC, Lawrence Berkeley National Laboratory, Berkeley, California 94720, USA\\
\texttt{luca.celotti@usherbrooke.ca}\quad\texttt{ermalrrapaj@lbl.gov}
}

\begin{document}

\maketitle

\begin{abstract}
    The softmax attention mechanism has emerged as a noteworthy development in the field of Artificial Intelligence research, building on the successes of Transformer-based architectures. However, their ever increasing sizes necessitate ever increasing computational memory, that limits their usage. We propose KgV, a sigmoid gating mechanism that, in conjunction with softmax attention, significantly boosts performance without increasing architecture size. To amend the size requirements, we leverage Tensor Chains to identify and prune the excess parameters. We find that such excess resides primarily within the embedding layer, and not in the output linear layer. To further improve embedding and significantly reduce parameters, we introduce H-SoftPOS, a hierarchical embedding layer which simultaneously enhances performance. Remarkably, on the WMT14 English-German validation set, our approach yields a threefold reduction in perplexity, surpassing the current state-of-the-art, while reducing parameter counts also by a factor of 3. When we further reduce the number of parameters up to sevenfold, we can still achieve a 21\% decrease in perplexity with respect to the baseline Transformer. To understand generalization capabilities, we conduct experiments on the 7 language pairs of the WMT17 dataset. Our method outperforms existing techniques in terms of test loss while simultaneously halving the number of parameters. Moreover, we observe a 70 times reduction in variance with respect to the prior state-of-the-art. In conclusion, our proposed method yields significant improvements in performance and much lower memory cost. We call the resulting architecture Anthe.
\end{abstract}
\section{Introduction}


The Transformer architecture~\citep{vaswani2017attention} has been a catalyst for major breakthroughs in Artificial Intelligence, leading to outstanding performance on a wide range of tasks, including language modeling \citep{brown2020language}, translation \citep{vaswani2017attention}, speech recognition \citep{radford2022robust}, and protein folding \citep{jumper2021highly}, among others.
Since their inception, Transformer-based architectures have become increasingly wider~\citep{brown2020language} and deeper~\citep{wang2022deepnet}, leading to a massive increase in the number of parameters. For instance, ChatGPT-3 has 175 billion parameters~\citep{Radford2019,brown2020language}, surpassing the number of parameters of previous models by orders of magnitude.  To address the huge computational demands of such architectures, especially in handling long sequences, researchers have proposed several approximate attention mechanisms, such as sparse-approximation~\citep{kitaev2020reformer, roy2021efficient}, low-rank approximation~\citep{wang2020linformer, katharopoulos2020transformers, choromanski2020rethinking}, their combination~\citep{beltagy2020longformer, zaheer2020bigbird, manzilbigbird}, and I/O optimization techniques for additional speed-up~\citep{dao2022}. However, not enough attention has been given to the efficient use of limited parameters, and it often seems that a reduction in parameters has to come with a degraded performance \citep{sanh2019distilbert}.

In this article we propose a novel gating mechanism placed before the softmax attention that significantly improves performance, as evidenced by our experimental findings. 
Additionally, we demonstrate that removing weight-sharing between the output projection and both embeddings of the encoder and decoder can further improve performance, at the cost of increasing the number of parameters by 43\%. 
To mitigate this increase in parameters without compromising accuracy, we introduce two techniques: Hierarchical Soft Part of Speech (H-SoftPOS) and Tensor Chain (TC).
H-SoftPOS is based on the observation that language elements, such as sub-words, words or sentences, can have a very limited set of functional roles, and each of them can adapt its role according to the context, which requires a soft aspect.
Therefore, we propose a method to assign a learnable Part of Speech (SoftPOS) to each subword, which helps improve performance while decreasing the number of parameters in the embedding.
Additionally, TC allows us to represent a large matrix as a tensor product contraction, drastically reducing the number of learnable parameters in the architecture. It was originally proposed in physics to characterize the short-range entanglement in one-dimensional quantum systems~\citep{Verstraete2004,Pirvu2010}, and has since found many applications in physics and more recently in deep learning~\citep{Gao2020}. We name the resulting architecture the \textit{Anthe} for \textit{G\textbf{a}tes, a\textbf{n}d \textbf{T}C and \textbf{H}ierarchical SoftPOS for Att\textbf{e}ntion}.
In summary, we design the \textit{Anthe}, a slim architecture that improves performance over the Transformer and reduces the number of parameters.   

Our contributions are:

\begin{itemize}[noitemsep,nolistsep]
    \item we introduce the KgV, a gating mechanism between values and keys in the attention paradigm in Sec.~\ref{sec:kgv};
    \item we introduce Soft Part of Speech to reduce embedding parameters without loss in performance by accounting for the limited functionality word sets play in speech, in Sec.~\ref{sec:softpos};
    \item we introduce the TC to represent any matrix as a product of small tensors to drastically reduce the amount of trainable parameters in Sec.~\ref{sec:tensorchain};
    \item we report improvements of \textit{Anthe} over Transformer first on English-German language translation, and then on other seven language pairs in Sec.~\ref{sec:results}. 
\end{itemize}

\section{Efficiency: higher performance with fewer parameters}

In this section, we present our novel techniques for achieving higher performance with fewer parameters by changing the Transformer setup with our contributions.
Our findings suggest that having independent weights for the embedding matrices and output projection significantly improves performance, contrary to prior work. Additionally, we replace the feed-forward layer with the GEGLU layer 
\citep{shazeer2020glu, chowdhery2022palm, lin2022survey}, 
as we observe a small but statistically significant improvement in performance.
In the following, we proceed to propose KgV to improve performance, and H-SoftPOS and TC to reduce the number of parameters.

\begin{figure}
    \centering
    \begin{minipage}{.4\textwidth}
        \subfloat[KgV]{\includegraphics[width=\textwidth]{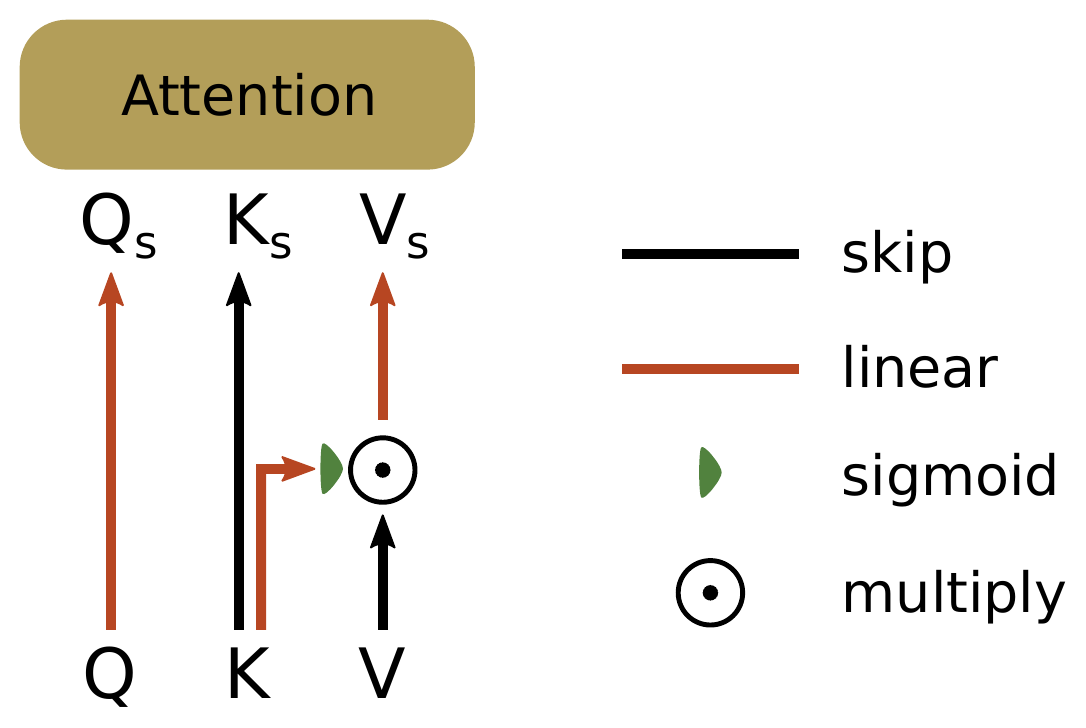}}
        \hspace{0.5cm}
        \subfloat[Tensor Chain (TC)]{\includegraphics[width=\textwidth]{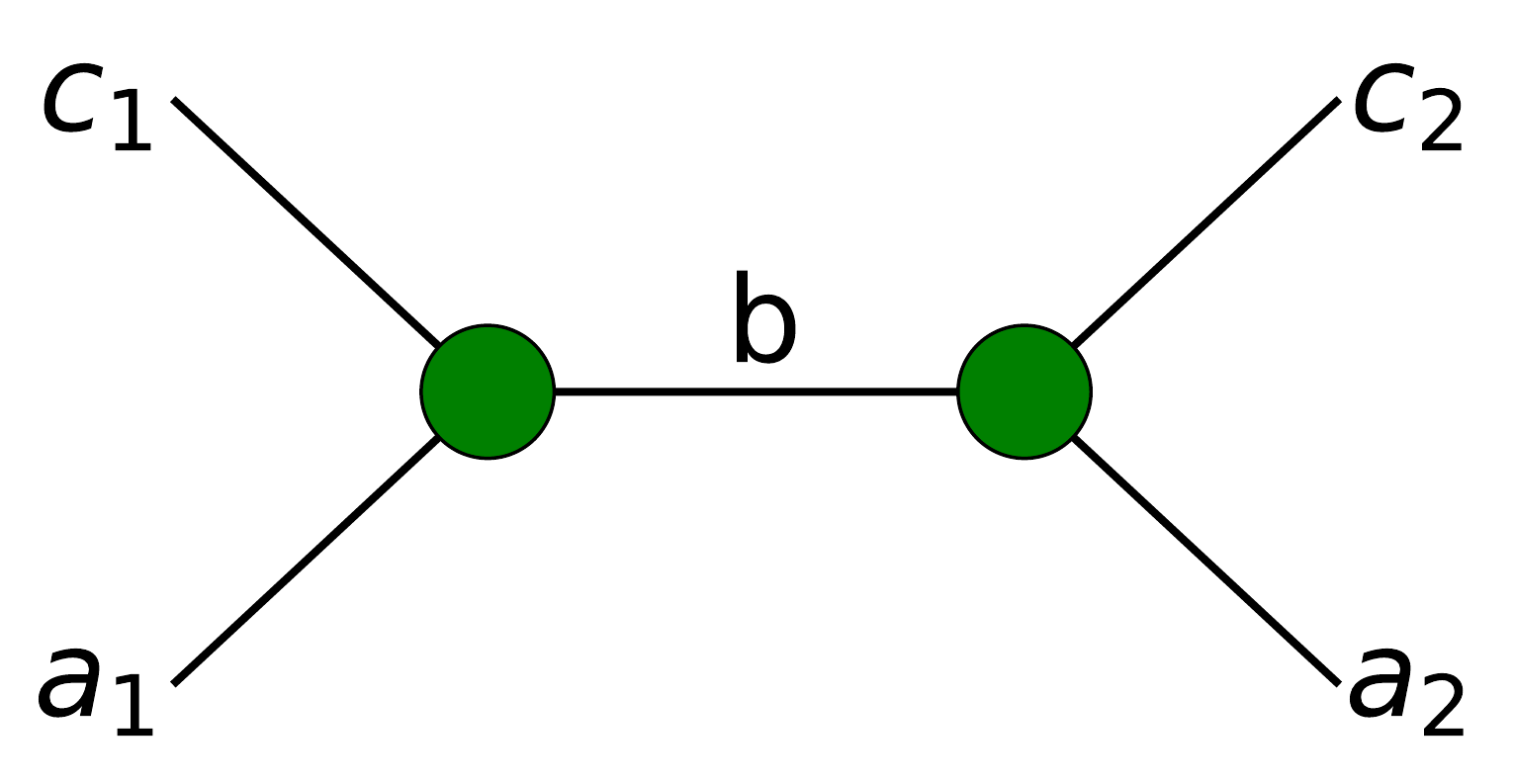}}
    \end{minipage}
    \hspace{0.5cm}
    \begin{minipage}{.4\textwidth}
        \subfloat[H-SoftPOS]{\includegraphics[width=\textwidth]{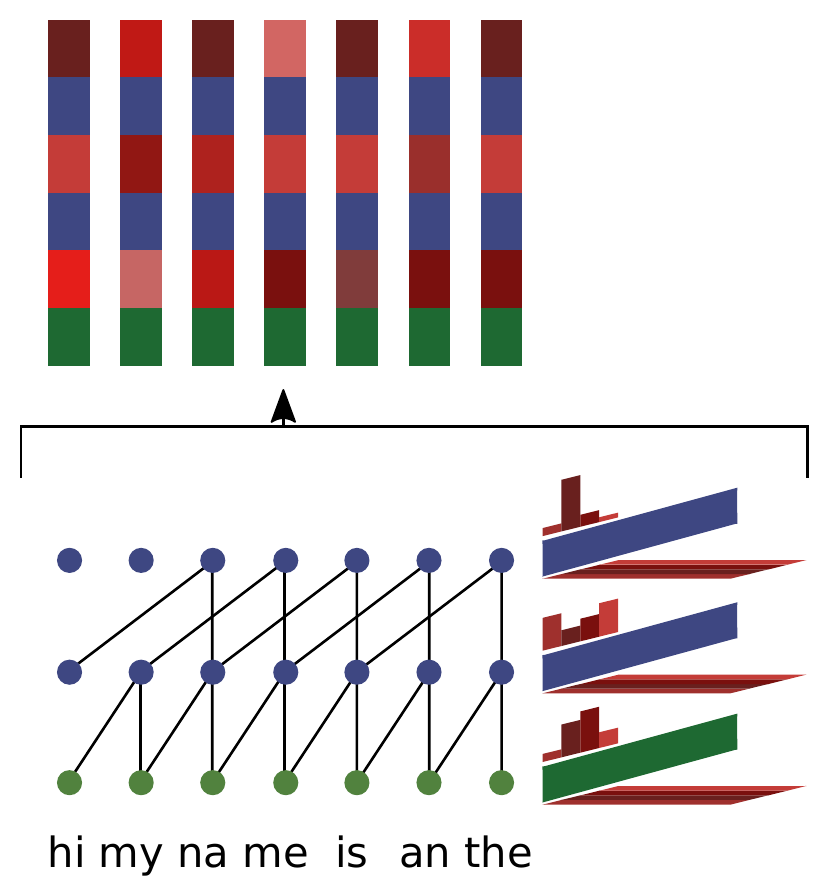}}
    \end{minipage}
    \hspace{0.5cm}
    \caption{\textbf{KgV, H-SoftPOS and TC.} (a) KgV uses the Key tensor to gate the Value tensor, before feeding KQV to the attention. (b) Length two Tensor Chain (TC) with bond dimension $b$ and external dimensions $N_a=a_1\cdot a_2,\ N_c = c_1\cdot c_2$. (c) H-SoftPOS starts with a smaller matrix embedding (green) that is concatenated with hierarchical temporal convolutions (blue) and SoftPOS of each hierarchical level (red). }
    \label{fig:architecture_layout}
\end{figure}

\subsection{Improving performance through gating}
\label{sec:kgv}

Gating mechanisms have been widely used in recurrent neural networks to avoid issues like gradient explosion \citep{hochreiter1997long}. The LSTM, GRU, Neural Turing Machine, Differentiable Neural Computer, and Mogrifier LSTM are well-known examples of models that employ gating mechanisms \citep{hochreiter1997long, chung2015gated, graves2014neural, graves2016hybrid, melismogrifier}.
In fact, the latter closed the gap in language modeling tasks between LSTM and Transformer. They observed a reduced variance in the LSTM gates when an additional external gating system was introduced, pointing at a stabilizing effect.
In the context of the Transformer architecture, queries, keys and values, are mapped linearly before applying the softmax attention, such that  $Q_s = W_QQ, K_s =W_KK, V_s = W_VV$.
In this work, for convenience of notation, we refer to $\{W_q, W_K, W_V\}$ as pre-attention, or \textit{patt}.
Inspired by the many successes of gates in the past, we propose a novel gating mechanism before the attention
\begin{equation}
\begin{aligned}
    V_s =& \ W_VV\sigma(W_KK)\\
    Q_s =& \ W_QQ \\   
    K_s =& \ K  \\
    Attention(Q_s,K_s,V_s) =& \ softmax\Big(\frac{Q_sK_s^T}{\sqrt{d_{model}/d_h}}\Big)V_s
\end{aligned}
\end{equation}
\noindent where $\sigma$ is the sigmoid activation function,  $d_{model}$ is the width of the architecture and $d_h$ the number of heads in the multi-head attention. Remarkably, it does not change the number of parameters with respect to the original linear map. In addition, we also tested all other gate combinations as shown in Tab.~\ref{tab:gates}, which resulted in less effective improvements.

\subsection{Reducing the number of parameters without compromising accuracy}

Compressing deep neural networks to reduce their number of learnable parameters while maintaining their prediction capabilities is a challenging and important problem, especially given the rapid increase in size of recent Transformer-based architectures. Such compression techniques can also help lower the risk of overfitting. The linear transformations used in fully connected, convolutional, and attention layers, contain the majority of the learnable parameters to be trained and stored. In particular, the embedding layers and output projection contribute significantly to the total number of parameters in the Transformer. Each of them is represented by a $d_{model}\times n_{vocab}$ matrix, resulting in three matrices of $16.3M$ parameters each, for standard choices of model width of $d_{model}=512$ and vocabulary size of $n_{vocab}=32K$.
To reduce the number of parameters, \cite{vaswani2017attention} proposed sharing the weights between these three layers. However, such weight sharing can lead to suboptimal results, as we demonstrate. Here, we propose two new approaches to significantly reduce the number of parameters while maintaining performance, H-SoftPOS and TC. These techniques allow us to keep the embeddings and output projection layers independent, while avoiding an explosion in the number of parameters.

\subsubsection{Hierarchical Soft Part of Speech (H-SoftPOS)}
\label{sec:softpos}

To reduce the number of parameters introduced by the embedding matrices, we propose a new approach called Hierarchical Soft Part of Speech (H-SoftPOS). The idea is that each sub-word can have a very limited set of functions in a word (e.g. prefix, suffix,  past tense of a verb, etc.). Similarly, each word can have a very limited set of functions in the sentence (e.g. verb, noun, adjective, etc.), and there's a limited set of subordinate clauses in a sentence (adverb, adjective, and noun clauses), and so on hierarchically. Since each of the elements can play different roles depending on the context, there is a soft aspect in the function assigned. We start with a much smaller embedding dimension and propose a method to assign a learnable Part of Speech (SoftPOS) to each subword. We use 1D convolutions to convert the sub-word embedding into word and sentence embeddings, and at each level we assign a hierarchical version of the SoftPOS idea. Finally, we concatenate the initial small embedding with the convolution levels and the SoftPOS representations, to have our full embedding representation. 
The matrix $W_{sp}\in\R^{n_{sp}\times d_{sp}}$  represents a finite set  $n_{sp}$ of possible POS functions, with dimension $d_{sp}$. We repeat the process at $l_{sp}$ hierarchical levels. If $S$ is the batch of sequences of integers that represent the input sentences, then
\begin{equation}
\begin{aligned}
    Embedding_{sp}(d_{model})(S) =& \ Concat \bigcup_{l=1}^{l_{sp}}\Big\{X_l, SoftPOS(X_l)\Big\}\\
    X_1=& \ Embedding(d_{emb})(S)\\
    X_l =& \ Conv1D(kernel=3, dilation=2^{l}, pad=causal)(X_{l-1}) \\
    SoftPOS(X_l) =& \ W_{sp}softmax(X_l[{:}{n_{sp}}])
\end{aligned}
\end{equation}
\noindent where in our implementation $l_{sp}=2$, $d_{sp}=\lfloor d_{model}/2l_{sp}\rfloor$ and $d_{emb}=d_{model} - (2l_{sp}-1)d_{sp}$. We use $X[{:}{n_{sp}}]$ to denote the first $n_{sp}$ elements of the vector. In our setting,  it results in an embedding of $d_{model}$ width, with four times fewer parameters and same performance with respect to the original version. The embedding $X_1$ has a matrix of size $d_{emb}\times n_{vocab}$ and sums a non-learnable cosine positional encoding \citep{vaswani2017attention}.

\subsubsection{Tensor Chain (TC) representation of a linear layer}
\label{sec:tensorchain}

The Tensor Chain (TC) representation, also known as the Matrix Product Operator (MPO), can be used to represent any linear transformation as a higher order tensor which is factorised into a sequential product of smaller tensors~\citep{Oseledets2011,Novikov2015}. The TC representation is a useful tool for quantum many-body systems, which are known to require extremely large number of parameters that could grow exponentially with the system size.
Although one might assume that the quality of the representation is degraded when the number of parameters is reduced, many applications in Condensed Matter Physics and Quantum Information have shown that efficient and faithful representations are possible~\citep{White1992,Vidal2033}. In the context of Deep Learning, \cite{Gao2020} replaced the linear layers with TCs in LeNet-5~\citep{Lecun1998}, VGG~\citep{Simonyan2015}, ResNet~\citep{He2015}, and DenseNet~\citep{Huang2016}, 
without any loss of prediction accuracy. 
A weight matrix $W_{N_a,N_c}$ of size $\R^{N_a\times N_c}$ can be decomposed as follows
\begin{equation}
W_{N_a,N_c} = \text{Reshape}\left\{ \text{Tr}_b\left[w^{(1)}_{a_1, b, c_1}\left(\prod_{i=2}^{n-1} w^{(i)}_{a_i, b,b,c_i} \right) w^{(n)}_{a_n, b, c_n}\right], N_a \times N_c\right\},
\end{equation}
where, $N_a=\prod_{i=1}^n a_i,\ N_b=\prod_{i=1}^n b_i$, $w_{abc}, w_{abbc}$ are tensors of order three and four in the chain and $n$ is the length of the chain. The trace is taken over the common index $b$, typically referred to as the bond. This internal index can vary between consecutive tensors, but we chose it to be the same for simplicity. It can be implemented through the well known einsum function, and here we provide a TC of length 3 as illustration,
\begin{equation}
 \text{einsum}\Big(a_1b_1c_1,\, a_2b_1b_2c_2, \, a_3b_2c_3\rightarrow a_1a_2a_3c_1c_2c_3,\left[w^{(1)}_{a_1, b_1, c_1},\,  w^{(2)}_{a_2, b_1,b_2,c_3},\,  w^{(3)}_{a_3, b_2,c_3}\right] \Big).
\end{equation}
\noindent 
For optimal memory allocation, the full matrix representation is only implicit since one can reshape the input in TC format, perform tensor contraction, and reshape the output into the target size \citep{Novikov:2018}. 

As we apply TC to different parts of the network, we use the notation TC$_{\textit{where}:r}$ to indicate the use of TC in the location \textit{where}, to reduce the number of parameters by a factor $r$. We denote by \textit{emb}, when the TC is applied to the embedding, \textit{ff} when applied to the feed-forward or the GEGLU module, \textit{patt} when applied to the pre-attention linear layers, \textit{layer} when applied both to \textit{patt} and \textit{ff}, and \textit{output} when applied to the last linear layer that outputs the logits. The parameter $r$ is a reduction factor, equal to the ratio between the number of parameters in the full linear matrix, with respect to the TC version. The bond parameter is the solution to  the equation
$b(a_1 c_1 + a_n c_n) + b^2 \sum_{i=2}^{n-1} a_i b_i = r N_a N_c$, for a given selection of parameters ($a_i, c_i$). In our setup, the external integer dimensions $a_i$ and $c_i$ of each tensor are chosen so as to be close to the $n$-th root of $N_a$ and $N_c$.
In principle, TC could be used to reduce the number of parameters after training, by performing a singular value decomposition (SVD) of the original matrix and pruning the small singular values. Such post processing step would not reduce the memory cost during training, but could be beneficial at inference time. 
In this work we focus on using TC during training, and we use a length $n=2$ unless stated otherwise. 
If the parameter reduction has minimal effect on loss, it can be concluded that many of the parameters in the original linear layer were of no functional importance.


\section{Results}
\label{sec:results}

In this section we summarize the results of our experiments with an ablation study of the improvements introduced above on the language translation task from WMT14 English to German. Then, we proceed to confirm that the improvements persist when applied to 7 new language pairs from the WMT17. All our experiments were conducted in the same fashion: we train for a maximum of two days with an NVIDIA V100 GPU and a batch size of 32. All our experiments finished by early stopping on the validation loss with a patience of 10 epochs. Our baseline Transformer has a width of $d_{model} = 512$, a number of layers of $N=6$, a number of heads of $d_h=8$, a dropout probability of $p_{dropout} = 0.1$, and a width for the feed-forward layer of $d_{\textit{ff}}= 4d_{model}$. For simplicity we use the same tokenizer for all the languages, which is a byte-pair encoding \citep{gage1994new} with 32K sub-words, trained on the WikiText-103 dataset \citep{merity2016pointer}.  We use the Adam optimizer \citep{adam} with $\beta_1=0.9$, $\beta_2=0.98$, $\epsilon=1e^{-9}$, and a learning rate of $lr = 3.16e^{-5}$ without any learning rate schedule. The learning rate was chosen after a grid search for optimal performance of the Transformer architecture on the WMT14 development set. To account for statistical fluctuations within the limits imposed by our computational resources, every reported result is the mean and standard deviation over 4 separate training experiments with different seeds.
Notice that our Transformer implementation results in 60M parameters because we used 32K sub-words, while the original 37K sub-words for the English-German pair, results in 63M parameters, closer to the 65M reported in \citep{vaswani2017attention}. 

\subsection{Ablation study} 
\label{sec:ablation}

\begin{table}[h]
\centering
\begin{tabular}{lrcc}
\toprule
& params & dev PPL \\
\midrule
Anthe = B' + KgV + H-SoftPOS + TC$_{\text{\textit{ff}:.005, \textit{patt}:.07}}$ & \textbf{\underline{30M}} & $\textbf{3.5674} \pm \textbf{0.0130}$  \\
B' + KgV + H-SoftPOS + TC$_{\text{\textit{ff}:.1}}$ & 46M & $2.7690 \pm 0.0048$  \\
B' + KgV + H-SoftPOS & \textbf{68M} & $\textbf{\underline{1.2627}} \pm \textbf{0.0018}$  \\
B' + KgV +  TC$_{\text{\textit{emb}:.2}}$ & \textbf{67M} & $\textbf{1.4146} \pm \textbf{0.0799}$ \\
B' + KgV & 93M & $1.2642 \pm 0.0035$  \\
B' = B + GEGLU & 93M & $2.5665 \pm 0.0055$  \\
B = Transformer + no-shared embeddings & 93M & $2.5987 \pm 0.0155$  \\ 
Transformer + shared embeddings & 60M &  $ 3.8245 \pm 0.0670$  \\
Transformer 512 \citep{vaswani2017attention} & 65M & 4.66 \\
Transformer 1024 \citep{vaswani2017attention} & 213M & 4.33 \\
\midrule
\bottomrule\\
\end{tabular}
\caption{\textbf{Anthe on the English-to-German translation development set WMT14.} We use the same hyper-parameters as \citep{vaswani2017attention}, for $d_{model}=512$, and we report at the bottom their two best results for $d_{model}=512, 1024$. Our KgV results in better performance, while H-SoftPOS slightly improves performance while reducing significantly the number of parameters. TC drastically reduces the number of parameters while retaining a better performance than the Transformer. The decrease in number of parameters with respect of Transformer 1024 is sevenfold, while retaining an improvement in performance.}
\label{tab:ablation}
\end{table}

\begin{table}
\begin{minipage}{.45\textwidth}
\centering
\begin{tabular}{crc}
\toprule
TC length & params & dev PPL\\
\midrule
 2 & 33M & $3.5592 \pm 0.0096$\\
 4 & 29M & $3.7582 \pm 0.1157$\\
 3 & 33M & $3.9060 \pm 0.0137$\\
\midrule
\bottomrule \\
\end{tabular}
\caption{\textbf{TCs}. The Anthe variant that we use in this study is B' + KgV + H-SoftPOS + TC$_{\text{\textit{layer}:.1}}$. The length of the TC has an impact on performance, the shortest being the best.}
\label{tab:tts}
\centering
\begin{tabular}{lc}
\toprule
gate & dev PPL\\
\midrule
B' + KgV & $\textbf{1.2642} \pm 0.0035$\\
B' + KgQ & $1.2771 \pm 0.0012$ \\
B' + QgV & $ 2.5341 \pm 0.0058$ \\
B' & $2.5665 \pm 0.0055$  \\
B' + QgK & $2.6079 \pm 0.0066$ \\
B' + VgK & $2.6113 \pm 0.0091$ \\
B' + VgQ & $2.6315 \pm 0.0093$ \\
\midrule
\bottomrule \\
\end{tabular}
\caption{\textbf{Gatings.} KgV outperforms all the other combinations of gating mechanisms. }

\label{tab:gates}
\end{minipage}
\hspace{1cm}
\begin{minipage}{.45\textwidth}
\centering
\begin{tabular}{lrc}
\toprule
TC & params & dev PPL\\
\midrule
B' + KgV & 93M & $1.2642 \pm 0.0035$  \\
B' + KgV + TC$_{\textit{emb}:.8}$ & 86M & $1.3787 \pm 0.0048$\\
B' + KgV + TC$_{\textit{emb}:.2}$ & 67M & $1.4146 \pm 0.0799$\\
B' + KgV + TC$_{\textit{layer}:.2}$ & 61M & $4.4466 \pm 0.0134$\\
B' + KgV + TC$_{\textit{layer}:.8}$ & 85M & $4.5861 \pm 0.0157$\\
B' + KgV + TC$_{\textit{output}:.2}$ & 80M & $6.8445 \pm 0.0985$\\
B' + KgV + TC$_{\textit{output}:.8}$ & 89M & $9.7301 \pm 0.1794$\\
\midrule
\bottomrule\\ 
\end{tabular}
\caption{
\textbf{Finding the excess parameters.} Parameters in the linear output layer seem to be much more important, since when reduced by 20\% and 80\%, it results in the strongest decrease in performance. However, decreasing the number of learnable parameters in the embedding has less of an impact on the performance. Notice the very non linear dependence with $r$.}
\label{tab:excess}
\end{minipage}
\end{table}

\begin{table}[h]
\centering
\begin{tabular}{lrc}
    \toprule
    gate & params & dev PPL\\
    \midrule
    Anthe  + no \textit{patt} & 29M & $\textbf{3.2351} \pm \textbf{0.0186}$\\
    Anthe  & 30M & $3.5674 \pm 0.0130$\\
    Anthe  + no \textit{patt} + no \textit{ff} & 29M & $4.3444 \pm 0.0359$ \\
    Anthe  + no \textit{ff} & 30M & $ 4.9069 \pm 0.0266$ \\
    \midrule
    \bottomrule \\
\end{tabular}
\caption{\textbf{Removing or TC?} Removing completely the pre-attention linear layers improves performance with respect to using TC on them, while removing the linear layers in the GEGLU worsens performance with respect to TC.}
\label{tab:remove}
\end{table}

\begin{table}[h]
\centering
\begin{tabular}{lcccc}
\toprule
& & CS/EN & DE/EN & FI/EN \\
dataset size & & 1.6G & 8.1G & 3.9G \\
& params & test PPL & test PPL & test PPL  \\
\midrule
Anthe + no \textit{patt}
  & 29M & $\textbf{4.6167} \pm \textbf{0.6501}$ & $\textbf{3.7891} \pm \textbf{0.0444}$ & $\textbf{4.2269} \pm \textbf{0.1164}$  \\
Anthe  & 30M & $\textbf{4.5259} \pm \textbf{0.5028}$ & $3.9822 \pm 0.0131$ & $\textbf{4.1558} \pm \textbf{0.0202}$ \\
Anthe$_{no TC}$  & 68M & $5.1874 \pm 1.1889$ & $5.3560 \pm 0.0480$ & $5.8806 \pm 0.1741$ \\
Transformer & 60M & $11.1995 \pm 4.4999$ & $6.2168 \pm 0.5581$ & $6.8927 \pm 1.4090$\\  
\midrule
\bottomrule \\
\end{tabular}

\vspace{.5cm}

\centering
\addtolength{\leftskip} {-1cm}
\addtolength{\rightskip}{-1cm}
\resizebox{1.1\textwidth}{!}{%
\begin{tabular}{lrcccc}
\toprule
 & & LV/EN & RU/EN & TR/EN & ZH/EN\\
 & & 4.3G & 9G & 306M & 9G \\
& params & test PPL & test PPL & test PPL & test PPL  \\
\midrule
Anthe + no \textit{patt}
  & 29M & $9.4998 \pm 0.6706$ & $\textbf{4.2085} \pm \textbf{0.1171}$ & $4.1760 \pm 0.0125$  & $\textbf{6.3080} \pm \textbf{0.2609}$  \\
Anthe  & 30M & $\textbf{8.8340} \pm \textbf{0.1074}$ & $\textbf{4.2563} \pm \textbf{0.0838}$ & $\textbf{4.1473} \pm \textbf{0.0046}$ & $\textbf{6.4599} \pm \textbf{0.0456}$ \\
Anthe$_{no TC}$  & 68M & $\textbf{8.2546} \pm \textbf{0.1130}$ & $6.6637 \pm 0.1169$ & $4.1697 \pm 0.0190$ & $9.5552 \pm 0.3533$ \\
Transformer & 60M & $10.9181 \pm 1.9593$ & $8.2324 \pm 0.1390$ & $5.4590 \pm 0.0262$ & $15.3343 \pm 0.8857$\\ 
\midrule
\bottomrule \\
\end{tabular}
}
\caption{\textbf{Different Languages.} Test perplexity on all the WMT17 language pairs. The Anthe outperforms the Transformer in all the language pairs with just half the parameters, with improvements in perplexity up to a factor of two, e.g. CS/EN. We also report on experiments with complete removal of the pre-attention linear layers which does not significantly improve Anthe, but also does not cause degradation.}
\label{tab:testlangs}
\end{table}

We introduce our innovations sequentially in Tab.~\ref{tab:ablation}, on the WMT14 English-German dataset, validation split. For simplicity, we denote as a reference baseline B the Transformer without weight sharing, and as B' when we change in B the feed-forward module by a GEGLU module. We first see that removing the shared weights between encoder and decoder embedding and output projections, increases the performance at the cost of an increment of $33M$ parameters, a 43\%. 
Instead, the GEGLU module brings a minor but significant increase in performance without an increase in parameters. The gating mechanism provided by KgV drastically improves performance without any cost in terms of parameters. In experiments not reported in the table, we observe an improvement of the Transformer perplexity from $3.8245 \pm 0.0670$ to $3.4310 \pm 0.0179$ with only the addition of the KgV to the original Transformer with weight sharing between output linear layer and embeddings. H-SoftPOS comes with a minor improvement in performance but a significant improvement in parameters reduction. Finally, TC allows us to bring down the number of parameters well below the original Transformer while maintaining  a better performance. We note that drastic reductions of the number of parameters through TC eventually degrade the performance as shown in the table. We explored a wide range of $r$ values in our experiments, and we report the best combinations.

In contrast to the common practice of sharing the weights between embedding and output layer, it is interesting to remark that the performance suffers the most when the TC is applied to the linear layer at the output, while performance suffers the least when TC is applied to the embedding, as it can be seen in Tab.~\ref{tab:excess}. This suggests that the embedding has an excess amount of parameters that can be pruned, while the output linear layer might be more important than often assumed. Additionally, we show in Tab.~\ref{tab:gates} that KgV outperforms every other combination of gating mechanisms. Moreover, we find combinations that degrade perplexity; an indication that care must be taken on how gating is applied.
Also we explore different lengths of TC in Tab.~\ref{tab:tts}, and a length of 2 gives the best results. Finally, we see in Tab.~\ref{tab:remove} that 
completely removing pre-attention improves performance while the removal of \textit{ff} degrades it. However, as we have shown in Tab.~\ref{tab:ablation}, one can aggressively apply TC to \textit{ff} without negative effects.

\subsection{Multiple language translation pairs }
\label{sec:multiplelangs}

We consider all the 7 different language pairs from the WMT17 datasets in Tab.~\ref{tab:testlangs}. We early stop on the validation set and we report on the test set. Both the RU/EN and the ZH/EN pairs, exceeded the 9G in size, so we limited them to 9G to make better use of our limited resources.
We compare the Transformer with our architecture with the least parameters and improved perplexity, and the best one in terms of perplexity from Tab.~\ref{tab:ablation}.
Remarkably the small Anthe outperforms the Transformer in all language pairs, with only half its parameters, and reduces the variance in the results up to 70 times, see FI/EN pair.
In addition, completely removing \textit{patt} from Anthe generally causes a small improvement, apart from the LV/EN and TR/EN language pairs.

\section{Discussion and Conclusion}

Introducing KgV, a sigmoid gating mechanism, as well as H-SoftPOS, a hierarchical embedding layer, and TC, tensor chain representation, we were able to significantly reduce the number of parameters required while enhancing performance. We call Anthe the resulting architecture. Our experimental results on both the WMT14 English-German validation set and the WMT17 test set for seven language pairs, indicate that our proposed method outperforms current state-of-the-art in terms of perplexity, while reducing parameter counts by at least a factor of two. 

Our analysis has confirmed that an excess of parameters exists within the Transformer-based architectures, which we were able to identify using H-SoftPOS and TC. In fact, and contrary to common practice, the embedding layer can be significantly pruned without major losses in performance, while the output linear layer needs all of its learnable parameters. We also observe that the feed-forward layer can be pruned more than the pre-attention linear projections. Surprisingly, our Anthe has more than half of all its parameters in the linear readout layer.

In light of these findings, we believe that our approach holds great promise for further advancing the field of Artificial Intelligence research in language translation and perhaps in language modeling which could be studied in future works.



\bibliographystyle{abbrvnat}
\bibliography{sections/references}

\begin{thebibliography}{39}
\providecommand{\natexlab}[1]{#1}
\providecommand{\url}[1]{\texttt{#1}}
\expandafter\ifx\csname urlstyle\endcsname\relax
  \providecommand{\doi}[1]{doi: #1}\else
  \providecommand{\doi}{doi: \begingroup \urlstyle{rm}\Url}\fi

\bibitem[Beltagy et~al.(2020)Beltagy, Peters, and Cohan]{beltagy2020longformer}
I.~Beltagy, M.~E. Peters, and A.~Cohan.
\newblock Longformer: The long-document transformer.
\newblock \emph{CoRR}, abs/2004.05150, 2020.
\newblock URL \url{https://arxiv.org/abs/2004.05150}.

\bibitem[Brown et~al.(2020)Brown, Mann, Ryder, Subbiah, Kaplan, Dhariwal,
  Neelakantan, Shyam, Sastry, Askell, Agarwal, Herbert-Voss, Krueger, Henighan,
  Child, Ramesh, Ziegler, Wu, Winter, Hesse, Chen, Sigler, Litwin, Gray, Chess,
  Clark, Berner, McCandlish, Radford, Sutskever, and Amodei]{brown2020language}
T.~Brown, B.~Mann, N.~Ryder, M.~Subbiah, J.~D. Kaplan, P.~Dhariwal,
  A.~Neelakantan, P.~Shyam, G.~Sastry, A.~Askell, S.~Agarwal, A.~Herbert-Voss,
  G.~Krueger, T.~Henighan, R.~Child, A.~Ramesh, D.~Ziegler, J.~Wu, C.~Winter,
  C.~Hesse, M.~Chen, E.~Sigler, M.~Litwin, S.~Gray, B.~Chess, J.~Clark,
  C.~Berner, S.~McCandlish, A.~Radford, I.~Sutskever, and D.~Amodei.
\newblock Language models are few-shot learners.
\newblock In H.~Larochelle, M.~Ranzato, R.~Hadsell, M.~Balcan, and H.~Lin,
  editors, \emph{Advances in Neural Information Processing Systems}, volume~33,
  pages 1877--1901. Curran Associates, Inc., 2020.
\newblock URL
  \url{https://proceedings.neurips.cc/paper_files/paper/2020/file/1457c0d6bfcb4967418bfb8ac142f64a-Paper.pdf}.

\bibitem[Chen et~al.(2021)Chen, Dao, Winsor, Song, Rudra, and
  R{\'{e}}]{zaheer2020bigbird}
B.~Chen, T.~Dao, E.~Winsor, Z.~Song, A.~Rudra, and C.~R{\'{e}}.
\newblock Scatterbrain: Unifying sparse and low-rank attention approximation.
\newblock \emph{CoRR}, abs/2110.15343, 2021.
\newblock URL \url{https://arxiv.org/abs/2110.15343}.

\bibitem[Choromanski et~al.(2020)Choromanski, Likhosherstov, Dohan, Song, Gane,
  Sarl{\'{o}}s, Hawkins, Davis, Mohiuddin, Kaiser, Belanger, Colwell, and
  Weller]{wang2020linformer}
K.~Choromanski, V.~Likhosherstov, D.~Dohan, X.~Song, A.~Gane, T.~Sarl{\'{o}}s,
  P.~Hawkins, J.~Davis, A.~Mohiuddin, L.~Kaiser, D.~Belanger, L.~J. Colwell,
  and A.~Weller.
\newblock Rethinking attention with performers.
\newblock \emph{CoRR}, abs/2009.14794, 2020.
\newblock URL \url{https://arxiv.org/abs/2009.14794}.

\bibitem[Chowdhery et~al.(2022)Chowdhery, Narang, Devlin, Bosma, Mishra,
  Roberts, Barham, Chung, Sutton, Gehrmann, et~al.]{chowdhery2022palm}
A.~Chowdhery, S.~Narang, J.~Devlin, M.~Bosma, G.~Mishra, A.~Roberts, P.~Barham,
  H.~W. Chung, C.~Sutton, S.~Gehrmann, et~al.
\newblock Palm: Scaling language modeling with pathways.
\newblock \emph{arXiv preprint arXiv:2204.02311}, 2022.

\bibitem[Chung et~al.(2015)Chung, Gulcehre, Cho, and Bengio]{chung2015gated}
J.~Chung, C.~Gulcehre, K.~Cho, and Y.~Bengio.
\newblock Gated feedback recurrent neural networks.
\newblock In \emph{International conference on machine learning}, pages
  2067--2075. PMLR, 2015.

\bibitem[{Dao} et~al.(2022){Dao}, {Fu}, {Ermon}, {Rudra}, and
  {R{\'e}}]{dao2022}
T.~{Dao}, D.~Y. {Fu}, S.~{Ermon}, A.~{Rudra}, and C.~{R{\'e}}.
\newblock {FlashAttention: Fast and Memory-Efficient Exact Attention with
  IO-Awareness}.
\newblock \emph{arXiv e-prints}, art. arXiv:2205.14135, May 2022.
\newblock \doi{10.48550/arXiv.2205.14135}.

\bibitem[Gage(1994)]{gage1994new}
P.~Gage.
\newblock A new algorithm for data compression.
\newblock \emph{C Users Journal}, 12\penalty0 (2):\penalty0 23--38, 1994.

\bibitem[Gao et~al.(2020)Gao, Cheng, He, Xie, Zhao, Lu, and Xiang]{Gao2020}
Z.-F. Gao, S.~Cheng, R.-Q. He, Z.~Y. Xie, H.-H. Zhao, Z.-Y. Lu, and T.~Xiang.
\newblock Compressing deep neural networks by matrix product operators.
\newblock \emph{Phys. Rev. Res.}, 2:\penalty0 023300, Jun 2020.
\newblock \doi{10.1103/PhysRevResearch.2.023300}.
\newblock URL \url{https://link.aps.org/doi/10.1103/PhysRevResearch.2.023300}.

\bibitem[Graves et~al.(2014)Graves, Wayne, and Danihelka]{graves2014neural}
A.~Graves, G.~Wayne, and I.~Danihelka.
\newblock Neural turing machines.
\newblock \emph{arXiv preprint arXiv:1410.5401}, 2014.

\bibitem[Graves et~al.(2016)Graves, Wayne, Reynolds, Harley, Danihelka,
  Grabska-Barwi{\'n}ska, Colmenarejo, Grefenstette, Ramalho, Agapiou,
  et~al.]{graves2016hybrid}
A.~Graves, G.~Wayne, M.~Reynolds, T.~Harley, I.~Danihelka,
  A.~Grabska-Barwi{\'n}ska, S.~G. Colmenarejo, E.~Grefenstette, T.~Ramalho,
  J.~Agapiou, et~al.
\newblock Hybrid computing using a neural network with dynamic external memory.
\newblock \emph{Nature}, 538\penalty0 (7626):\penalty0 471--476, 2016.

\bibitem[He et~al.(2015)He, Zhang, Ren, and Sun]{He2015}
K.~He, X.~Zhang, S.~Ren, and J.~Sun.
\newblock Deep residual learning for image recognition.
\newblock \emph{CoRR}, abs/1512.03385, 2015.
\newblock URL \url{http://arxiv.org/abs/1512.03385}.

\bibitem[Hochreiter and Schmidhuber(1997)]{hochreiter1997long}
S.~Hochreiter and J.~Schmidhuber.
\newblock Long short-term memory.
\newblock \emph{Neural computation}, 9\penalty0 (8):\penalty0 1735--1780, 1997.

\bibitem[Huang et~al.(2016)Huang, Liu, and Weinberger]{Huang2016}
G.~Huang, Z.~Liu, and K.~Q. Weinberger.
\newblock Densely connected convolutional networks.
\newblock \emph{CoRR}, abs/1608.06993, 2016.
\newblock URL \url{http://arxiv.org/abs/1608.06993}.

\bibitem[Jumper et~al.(2021)Jumper, Evans, Pritzel, Green, Figurnov,
  Ronneberger, Tunyasuvunakool, Bates, {\v{Z}}{\'\i}dek, Potapenko,
  et~al.]{jumper2021highly}
J.~Jumper, R.~Evans, A.~Pritzel, T.~Green, M.~Figurnov, O.~Ronneberger,
  K.~Tunyasuvunakool, R.~Bates, A.~{\v{Z}}{\'\i}dek, A.~Potapenko, et~al.
\newblock Highly accurate protein structure prediction with alphafold.
\newblock \emph{Nature}, 596\penalty0 (7873):\penalty0 583--589, 2021.

\bibitem[Katharopoulos et~al.(2020)Katharopoulos, Vyas, Pappas, and
  Fleuret]{katharopoulos2020transformers}
A.~Katharopoulos, A.~Vyas, N.~Pappas, and F.~Fleuret.
\newblock Transformers are rnns: Fast autoregressive transformers with linear
  attention.
\newblock \emph{CoRR}, abs/2006.16236, 2020.
\newblock URL \url{https://arxiv.org/abs/2006.16236}.

\bibitem[Kingma and Ba(2015)]{adam}
D.~P. Kingma and J.~Ba.
\newblock Adam: {A} method for stochastic optimization.
\newblock In Y.~Bengio and Y.~LeCun, editors, \emph{3rd International
  Conference on Learning Representations, {ICLR} 2015, San Diego, CA, USA, May
  7-9, 2015, Conference Track Proceedings}, 2015.
\newblock URL \url{http://arxiv.org/abs/1412.6980}.

\bibitem[Kitaev et~al.(2020)Kitaev, Kaiser, and Levskaya]{kitaev2020reformer}
N.~Kitaev, L.~Kaiser, and A.~Levskaya.
\newblock Reformer: The efficient transformer.
\newblock In \emph{8th International Conference on Learning Representations,
  {ICLR} 2020, Addis Ababa, Ethiopia, April 26-30, 2020}. OpenReview.net, 2020.
\newblock URL \url{https://openreview.net/forum?id=rkgNKkHtvB}.

\bibitem[Lecun et~al.(1998)Lecun, Bottou, Bengio, and Haffner]{Lecun1998}
Y.~Lecun, L.~Bottou, Y.~Bengio, and P.~Haffner.
\newblock Gradient-based learning applied to document recognition.
\newblock \emph{Proceedings of the IEEE}, 86\penalty0 (11):\penalty0
  2278--2324, 1998.
\newblock \doi{10.1109/5.726791}.

\bibitem[Likhosherstov et~al.(2020)Likhosherstov, Choromanski, Davis, Song, and
  Weller]{roy2021efficient}
V.~Likhosherstov, K.~Choromanski, J.~Davis, X.~Song, and A.~Weller.
\newblock Sub-linear memory: How to make performers slim.
\newblock \emph{CoRR}, abs/2012.11346, 2020.
\newblock URL \url{https://arxiv.org/abs/2012.11346}.

\bibitem[Lin et~al.(2022)Lin, Wang, Liu, and Qiu]{lin2022survey}
T.~Lin, Y.~Wang, X.~Liu, and X.~Qiu.
\newblock A survey of transformers.
\newblock \emph{AI Open}, 2022.

\bibitem[Melis et~al.(2019)Melis, Ko{\v{c}}isk{\`y}, and
  Blunsom]{melismogrifier}
G.~Melis, T.~Ko{\v{c}}isk{\`y}, and P.~Blunsom.
\newblock Mogrifier lstm.
\newblock In \emph{International Conference on Learning Representations}, 2019.

\bibitem[Merity et~al.(2016)Merity, Xiong, Bradbury, and
  Socher]{merity2016pointer}
S.~Merity, C.~Xiong, J.~Bradbury, and R.~Socher.
\newblock Pointer sentinel mixture models, 2016.

\bibitem[Novikov et~al.(2015)Novikov, Podoprikhin, Osokin, and
  Vetrov]{Novikov2015}
A.~Novikov, D.~Podoprikhin, A.~Osokin, and D.~P. Vetrov.
\newblock Tensorizing neural networks.
\newblock In C.~Cortes, N.~Lawrence, D.~Lee, M.~Sugiyama, and R.~Garnett,
  editors, \emph{Advances in Neural Information Processing Systems}, volume~28.
  Curran Associates, Inc., 2015.
\newblock URL
  \url{https://proceedings.neurips.cc/paper/2015/file/6855456e2fe46a9d49d3d3af4f57443d-Paper.pdf}.

\bibitem[Novikov et~al.(2020)Novikov, Izmailov, Khrulkov, Figurnov, and
  Oseledets]{Novikov:2018}
A.~Novikov, P.~Izmailov, V.~Khrulkov, M.~Figurnov, and I.~Oseledets.
\newblock Tensor train decomposition on tensorflow (t3f).
\newblock \emph{The Journal of Machine Learning Research}, 21\penalty0
  (1):\penalty0 1105--1111, 2020.

\bibitem[Oseledets(2011)]{Oseledets2011}
I.~V. Oseledets.
\newblock Tensor-train decomposition.
\newblock \emph{SIAM Journal on Scientific Computing}, 33\penalty0
  (5):\penalty0 2295--2317, 2011.
\newblock \doi{10.1137/090752286}.
\newblock URL \url{https://doi.org/10.1137/090752286}.

\bibitem[Pirvu et~al.(2010)Pirvu, Murg, Cirac, and Verstraete]{Pirvu2010}
B.~Pirvu, V.~Murg, J.~I. Cirac, and F.~Verstraete.
\newblock Matrix product operator representations.
\newblock \emph{New Journal of Physics}, 12\penalty0 (2):\penalty0 025012, feb
  2010.
\newblock \doi{10.1088/1367-2630/12/2/025012}.
\newblock URL \url{https://dx.doi.org/10.1088/1367-2630/12/2/025012}.

\bibitem[Radford et~al.(2019)Radford, Wu, Child, Luan, Amodei, Sutskever,
  et~al.]{Radford2019}
A.~Radford, J.~Wu, R.~Child, D.~Luan, D.~Amodei, I.~Sutskever, et~al.
\newblock Language models are unsupervised multitask learners.
\newblock \emph{OpenAI blog}, 1\penalty0 (8):\penalty0 9, 2019.

\bibitem[Radford et~al.(2022)Radford, Kim, Xu, Brockman, McLeavey, and
  Sutskever]{radford2022robust}
A.~Radford, J.~W. Kim, T.~Xu, G.~Brockman, C.~McLeavey, and I.~Sutskever.
\newblock Robust speech recognition via large-scale weak supervision.
\newblock \emph{arXiv preprint arXiv:2212.04356}, 2022.

\bibitem[Sanh et~al.(2019)Sanh, Debut, Chaumond, and Wolf]{sanh2019distilbert}
V.~Sanh, L.~Debut, J.~Chaumond, and T.~Wolf.
\newblock Distilbert, a distilled version of bert: Smaller, faster, cheaper and
  lighter. arxiv 2019.
\newblock \emph{arXiv preprint arXiv:1910.01108}, 2019.

\bibitem[Shazeer(2020)]{shazeer2020glu}
N.~Shazeer.
\newblock Glu variants improve transformer.
\newblock \emph{arXiv preprint arXiv:2002.05202}, 2020.

\bibitem[Simonyan and Zisserman(2015)]{Simonyan2015}
K.~Simonyan and A.~Zisserman.
\newblock Very deep convolutional networks for large-scale image recognition.
\newblock In Y.~Bengio and Y.~LeCun, editors, \emph{3rd International
  Conference on Learning Representations, {ICLR} 2015, San Diego, CA, USA, May
  7-9, 2015, Conference Track Proceedings}, 2015.
\newblock URL \url{http://arxiv.org/abs/1409.1556}.

\bibitem[Vaswani et~al.(2017)Vaswani, Shazeer, Parmar, Uszkoreit, Jones, Gomez,
  Kaiser, and Polosukhin]{vaswani2017attention}
A.~Vaswani, N.~Shazeer, N.~Parmar, J.~Uszkoreit, L.~Jones, A.~N. Gomez,
  {\L}.~Kaiser, and I.~Polosukhin.
\newblock Attention is all you need.
\newblock In \emph{NeurIPS}, pages 5998--6008, 2017.

\bibitem[Verstraete et~al.(2004)Verstraete, Garc\'{\i}a-Ripoll, and
  Cirac]{Verstraete2004}
F.~Verstraete, J.~J. Garc\'{\i}a-Ripoll, and J.~I. Cirac.
\newblock Matrix product density operators: Simulation of finite-temperature
  and dissipative systems.
\newblock \emph{Phys. Rev. Lett.}, 93:\penalty0 207204, Nov 2004.
\newblock \doi{10.1103/PhysRevLett.93.207204}.
\newblock URL \url{https://link.aps.org/doi/10.1103/PhysRevLett.93.207204}.

\bibitem[Vidal(2003)]{Vidal2033}
G.~Vidal.
\newblock Efficient classical simulation of slightly entangled quantum
  computations.
\newblock \emph{Phys. Rev. Lett.}, 91:\penalty0 147902, Oct 2003.
\newblock \doi{10.1103/PhysRevLett.91.147902}.
\newblock URL \url{https://link.aps.org/doi/10.1103/PhysRevLett.91.147902}.

\bibitem[{Wang} et~al.(2022){Wang}, {Ma}, {Dong}, {Huang}, {Zhang}, and
  {Wei}]{wang2022deepnet}
H.~{Wang}, S.~{Ma}, L.~{Dong}, S.~{Huang}, D.~{Zhang}, and F.~{Wei}.
\newblock {DeepNet: Scaling Transformers to 1,000 Layers}.
\newblock \emph{arXiv e-prints}, art. arXiv:2203.00555, Mar. 2022.
\newblock \doi{10.48550/arXiv.2203.00555}.

\bibitem[Wang et~al.(2020)Wang, Li, Khabsa, Fang, and
  Ma]{choromanski2020rethinking}
S.~Wang, B.~Z. Li, M.~Khabsa, H.~Fang, and H.~Ma.
\newblock Linformer: Self-attention with linear complexity.
\newblock \emph{CoRR}, abs/2006.04768, 2020.
\newblock URL \url{https://arxiv.org/abs/2006.04768}.

\bibitem[White(1992)]{White1992}
S.~R. White.
\newblock Density matrix formulation for quantum renormalization groups.
\newblock \emph{Phys. Rev. Lett.}, 69:\penalty0 2863--2866, Nov 1992.
\newblock \doi{10.1103/PhysRevLett.69.2863}.
\newblock URL \url{https://link.aps.org/doi/10.1103/PhysRevLett.69.2863}.

\bibitem[Zaheer et~al.(2020)Zaheer, Guruganesh, Dubey, Ainslie, Alberti,
  Onta{\~{n}}{\'{o}}n, Pham, Ravula, Wang, Yang, and Ahmed]{manzilbigbird}
M.~Zaheer, G.~Guruganesh, A.~Dubey, J.~Ainslie, C.~Alberti,
  S.~Onta{\~{n}}{\'{o}}n, P.~Pham, A.~Ravula, Q.~Wang, L.~Yang, and A.~Ahmed.
\newblock Big bird: Transformers for longer sequences.
\newblock \emph{CoRR}, abs/2007.14062, 2020.
\newblock URL \url{https://arxiv.org/abs/2007.14062}.

\end{thebibliography}

\begin{appendix}

\end{appendix}

\end{document}